\documentclass[conference]{IEEEtran}
\IEEEoverridecommandlockouts

\usepackage{cite}
\usepackage{amsmath,amssymb,amsfonts}
\usepackage{algorithmic}
\usepackage{graphicx}
\usepackage{multirow}
\usepackage{textcomp}
\usepackage{xcolor}
\usepackage{booktabs}
\usepackage{bm}

\def\BibTeX{{\rm B\kern-.05em{\sc i\kern-.025em b}\kern-.08em
    T\kern-.1667em\lower.7ex\hbox{E}\kern-.125emX}}
\begin{document}

\title{Delta-ICM: Entropy Modeling with Delta Function for Learned Image Compression\\
}

\author{\IEEEauthorblockN{Takahiro Shindo}
\IEEEauthorblockA{\textit{Graduate School of FSE,}\\
\textit{Waseda University}\\
Tokyo, Japan \\
taka\_s0265@ruri.waseda.jp}
\and
\IEEEauthorblockN{Taiju Watanabe}
\IEEEauthorblockA{\textit{Graduate School of FSE,}\\
\textit{Waseda University}\\
Tokyo, Japan \\
lvpurin@fuji.waseda.jp}
\and
\IEEEauthorblockN{Yui Tatsumi}
\IEEEauthorblockA{\textit{School of FSE,}\\
\textit{Waseda University}\\
Tokyo, Japan \\
yui.t@fuji.waseda.jp}
\and
\IEEEauthorblockN{Hiroshi Watanabe}
\IEEEauthorblockA{\textit{Graduate School of FSE,}\\
\textit{Waseda University}\\
Tokyo, Japan \\
hiroshi.watanabe@waseda.jp}
}

\maketitle

\begin{abstract}
    Image Coding for Machines (ICM) is becoming more important as research in computer vision progresses.
    ICM is a vital research field that pursues the use of images for image recognition models, facilitating efficient image transmission and storage.
    Demand and required performance for recognition models are rapidly growing within consumers.
    To meet these needs, exchanging image data between consumer devices and cloud AI using ICM technology could be one possible solution.
    In ICM, various image compression methods have adopted Learned Image Compression (LIC).
    LIC includes an entropy model for estimating the bitrate of latent features, and the design of this model significantly affects its performance.
    Typically, LIC methods assume that the distribution of latent features follows a normal distribution.
    This assumption is effective for compressing images intended for human vision.
    However, employing an entropy model based on normal distribution is inefficient in ICM due to the limitation of image parts that require precise decoding.
    To address this, we propose Delta-ICM, which uses a probability distribution based on a delta function.
    Assuming the delta distribution as a distribution of latent features reduces the entropy of image portions unnecessary for machines.
    We compress the remaining portions using an entropy model based on normal distribution, similar to existing methods.
    Delta-ICM selects between the entropy model based on the delta distribution and the one based on the normal distribution for each latent feature.
    Our method outperforms existing ICM methods in image compression performance aimed at machines.
\end{abstract}


\section{Introduction}
Following the advancement of research in computer vision, image recognition models continues to improve in terms of performance.
Expectations for machine-based image analysis techniques grow along with the rapid expansion of utilizing image recognition models.
Since the use of recognition models by the general public is increasing, the challenge is to meet the demand for their use.
To adress this challenge, it is essential to enable high-speed data exchange between consumer devices and cloud AI.
In addition, image data compression for machines is required for efficient data use in smart devices.
In this context, image compression technology for these models is crucial, given the increasing quantity of image data needing recognition.
Previous image compression standards \cite{a1,a2,a3,a4} are designed for human vision and do not account for machine recognition.
Consequently, research on Image Coding for Machines (ICM) has intensified \cite{a5,a6,a7,a8,a9,a10,b3}.

Learned Image Compression (LIC) methods are widely used in image compression for ICM.
LIC is a neural network-based approach to image compression \cite{a11,a12,a13,b1}.
This approach employs an entropy model to estimate the bitrate of latent features, and the design of this model significantly affects the compression performance of LIC.
Typically, LIC assumes that the distribution of latent features follows a normal distribution when designing the entropy model \cite{a14,a15}.
This assumption is valid for accurate image decoding, which is particularly effective since much research on LIC focuses on image compression for human vision.
Numerous studies have demonstrated that LIC outperforms existing image compression standards \cite{a16,a17,a18,a19}, such as HEVC \cite{a3} and VVC \cite{a4}.

The LIC method with this entropy model is used in many studies on ICM.
Most of these studies attempt to improve compression performance by devising new loss functions to optimize LIC without altering the design of the entropy model.
For example, methods using task-loss \cite{a8,a9,a10,b2} and region-learning-loss \cite{a20,a21} have been successfully implemented in ICM.
ICM methods based on task-loss employ the accuracy of the image recognition task as a loss function for training LIC.
These methods enable effective image decoding for machines by learning image compression methods that maintain high recognition accuracy.
In the ICM method based on region-learning-loss, LIC learns the compression method for specific parts of the image.
This approach ensures that the critical parts required for image recognition are efficiently decoded.
These two loss function innovations contribute to improving the accuracy of ICM without redesigning the entropy model of LIC.
However, there is still room for improvement in the design of the entropy model for ICM.

We revise the design of the entropy model and propose a new ICM method, Delta-ICM.
By incorporating a probability distribution based on the delta function (delta distribution) into the entropy model, we achieve efficient image compression for machines.
Assuming the delta distribution for latent features reduces the entropy of parts of the image that are not needed for image recognition.
Other parts in the image are compressed using an entropy model based on a normal distribution, similar to existing methods.
Delta-ICM allows for the design of efficient entropy models by selecting either an entropy model based on the delta distribution or one based on the normal distribution for each latent feature in the image.
We demonstrate that our method offers better image compression performance for machines compared to existing ICM methods.
We verify the generality of our proposed method by applying it to object detection and instance segmentation tasks.

In summary, we make the following contributions.
\begin{itemize}
  \item We propose Delta-ICM, a method that enhances image compression for machines by assuming a delta distribution for the probability distribution of latent features.
  \item Our design includes an entropy model that utilizes both normal and delta distributions, allowing the LIC to be optimized by selecting the appropriate distribution for each latent feature.
  \item Delta-ICM outperforms other ICM methods in image compression performance for object detection and segmentation.
\end{itemize}

\begin{figure}[bt]
    \centerline{\includegraphics[width=0.95\columnwidth]{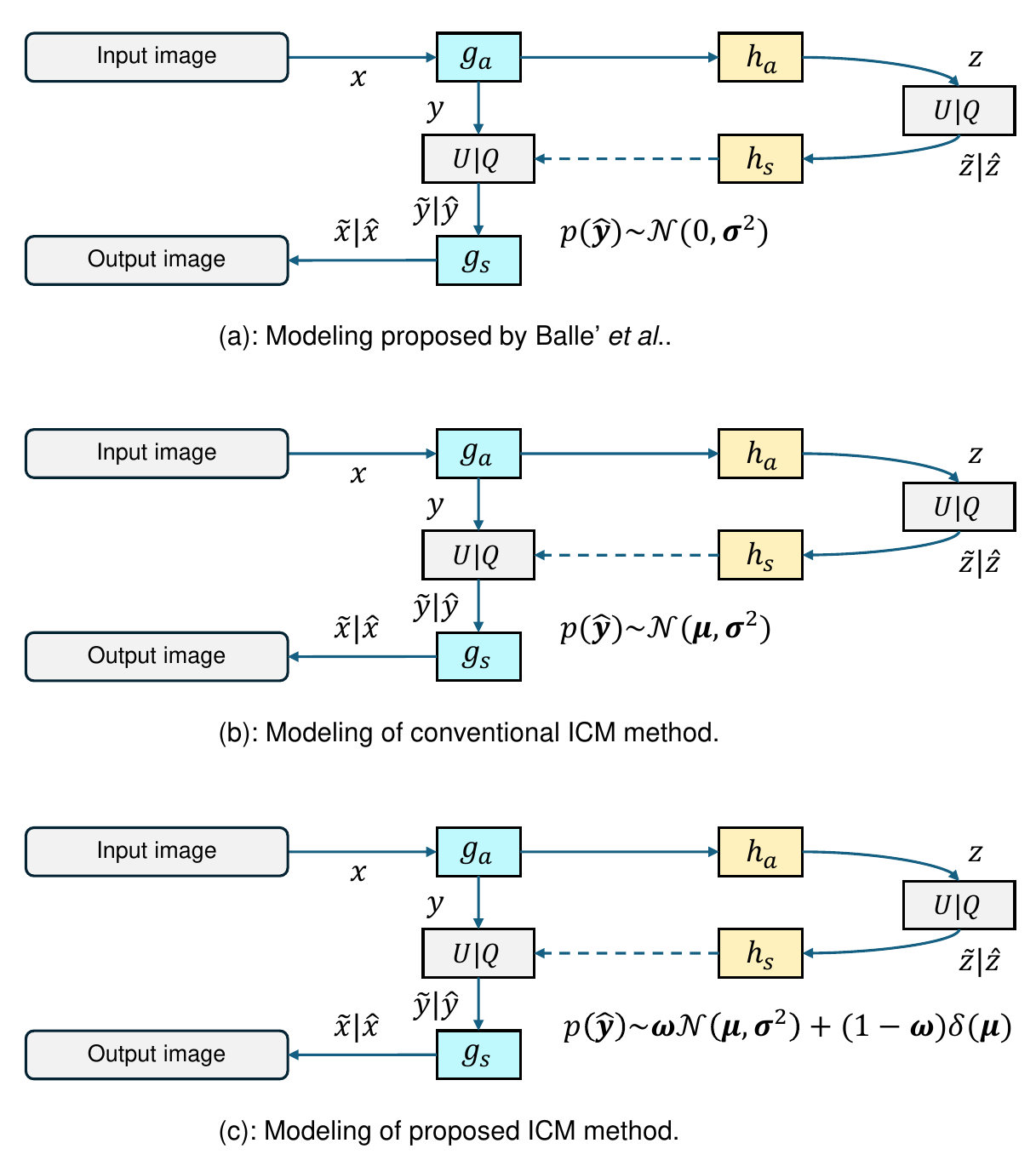}}
    \caption{Entropy model design. (a): LIC model proposed by Balle' \textit{et al}., (b): Conventional ICM method, using Gaussian distribution, (c): Proposed ICM method, using Gaussian and Delta distributions.}
    \label{fig:flow}
    \end{figure}
\begin{figure}[bt]
    \centerline{\includegraphics[width=0.95\columnwidth]{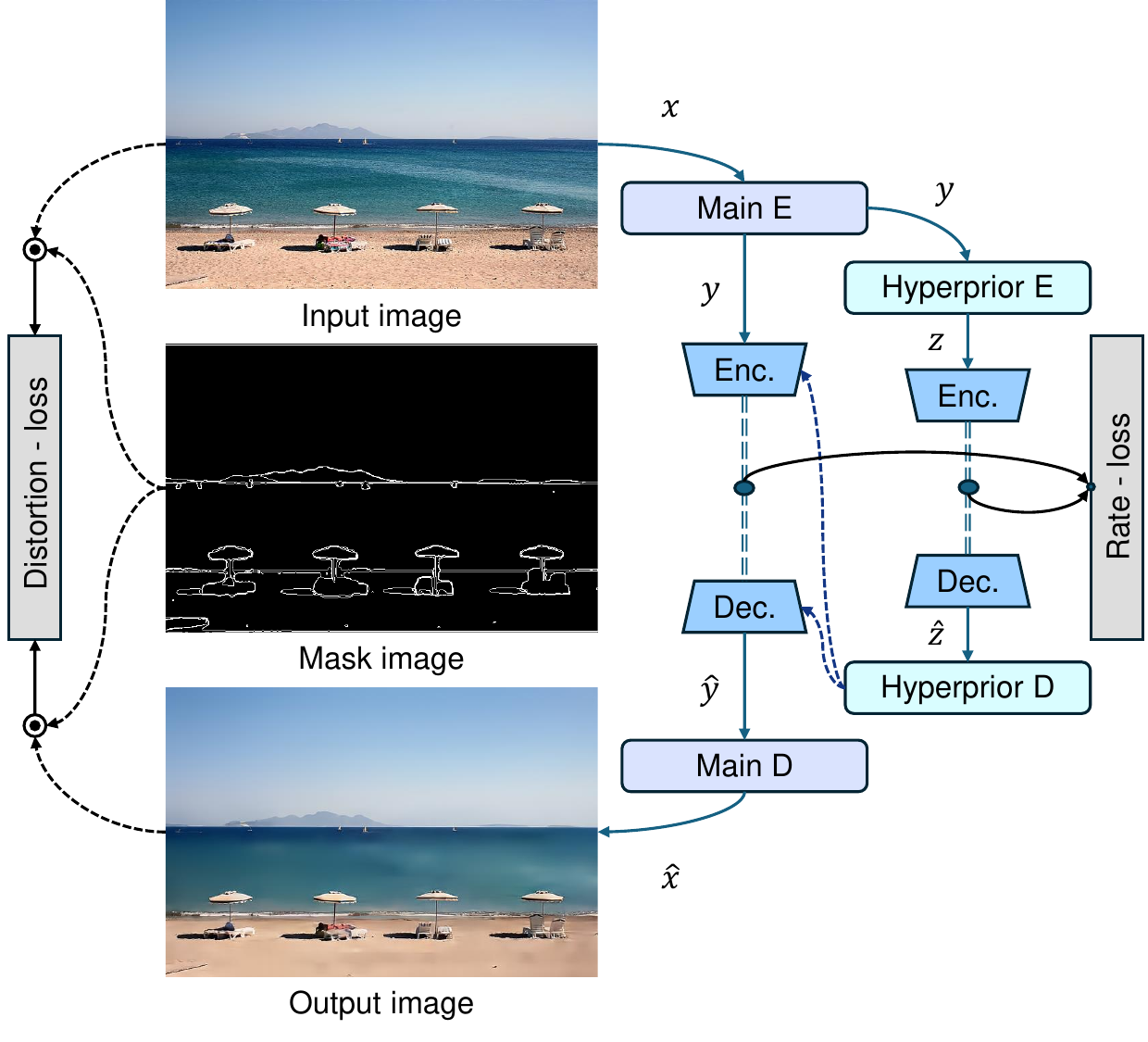}}
    \caption{Learning method for SA-ICM. Using the mask images, the learned image compression (LIC) model is trained to encode and decode specific regions of the image.}
    \label{fig:saicm}
    \end{figure}

\begin{figure*}[bt]
    \centerline{\includegraphics[width=1.9\columnwidth]{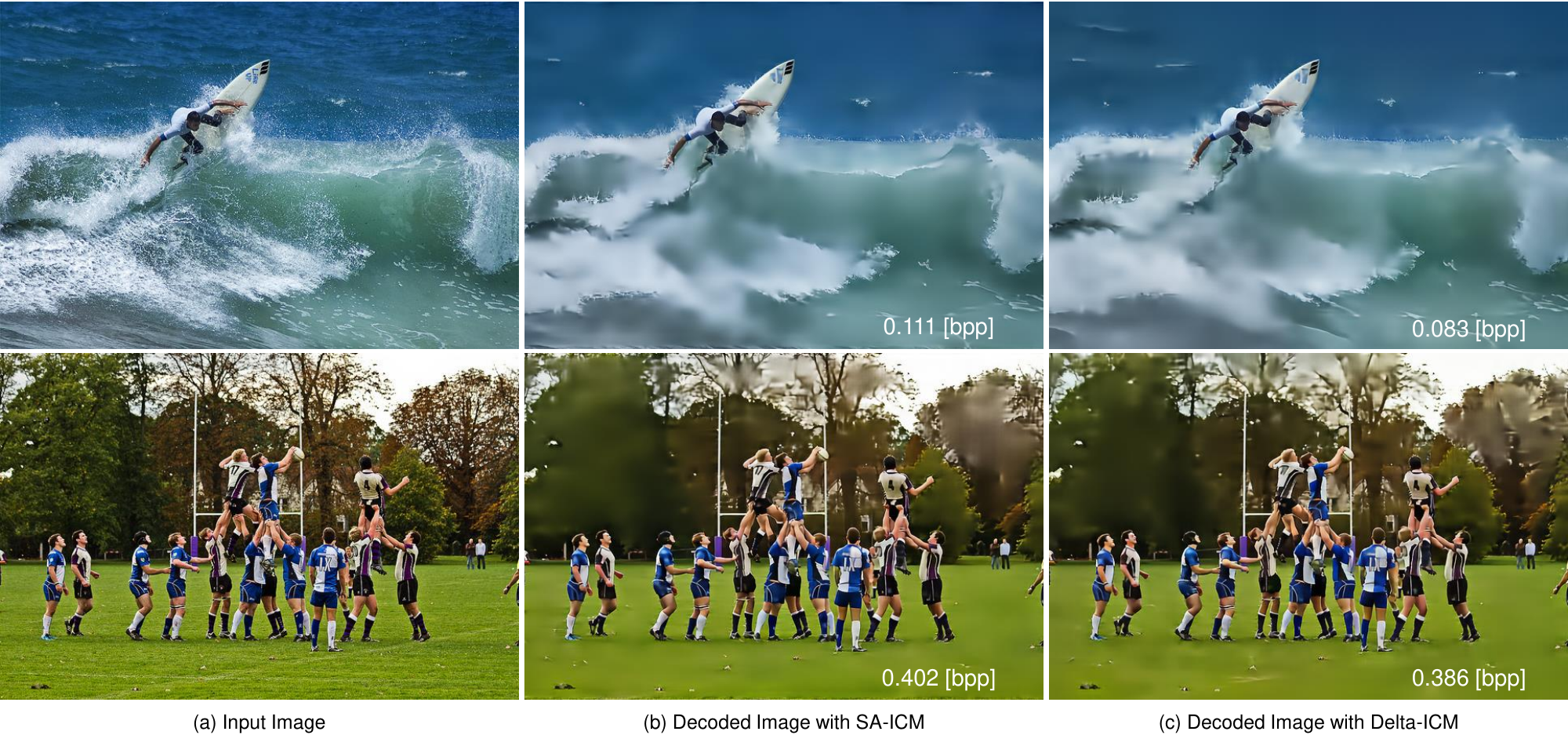}}
    \caption{Examples of coded images of the COCO2017 dataset and their bitrates. (a): Input image, (b): Decoded image by the conventional method (SA-ICM), (c): Decoded image by the proposed method (Delta-ICM).}
    \label{fig:exp}
    \end{figure*}

\section{Related Work}

\subsection{Entropy Model in Learned Image Compression}
Learned Image Compression (LIC) is a neural network-based image compression method \cite{a11,a12,a13}.
Many LIC methods have been proposed in recent years, and these typically include an entropy model to encode the features.
Most of these methods use the hyperpripr-based entropy model framework proposed by Balle' \textit{et al}.\cite{a14} to enable efficient coding.
This framework is shown in Fig.\ref{fig:flow}(a).
$\bm{y}$ is the latent feature obtained from the image $\bm{x}$ by the function $g_{a}$, represented in the neural network.
$\bm{z}$ is the feature that conditions $\bm{y}$, obtained from the latent feature $\bm{y}$ through the function $h_{a}$.
The probability distribution of the latent feature $\bm{y}$ conditioned on $\bm{z}$ is assumed to follow a normal distribution with mean $0$, that is,
\begin{equation}
  p_{\bm{\hat{y}}|\bm{\hat{z}}}(\bm{\hat{y}}|\bm{\hat{z}})\sim\mathcal{N}(0,\bm{\sigma}^2).
\end{equation}
In (1), $\bm{\sigma}$ is the value estimated from $\bm{\hat{z}}$ and represents the scale of the normal distribution.
The study by Minnen \textit{et al}.\cite{a15} extends this framework and assumes the probability distribution of the latent feature $\bm{y}$ as follows:
\begin{equation}
  p_{\bm{\hat{y}}|\bm{\hat{z}}}(\bm{\hat{y}}|\bm{\hat{z}})\sim\mathcal{N}(\bm{\mu},\bm{\sigma}^2).
\end{equation}
In (2), $\bm{\mu}$ and $\bm{\sigma}$ are both estimated from $\bm{\hat{z}}$ and represent the mean and scale of the normal distribution, respectively.
By assuming the normal distribution as the probability distribution of the feature $\bm{y}$, the entropy model is formulated as follows:
\begin{equation}
\begin{split}
  &p_{\bm{\hat{y}}|\bm{\hat{z}}}(\bm{\hat{y}}|\bm{\hat{z}})=\prod_{i}p_{\bm{\hat{y}}|\bm{\hat{z}}}(\hat{y}_{i}|\bm{\hat{z}})\\
  &p_{\bm{\hat{y}}|\bm{\hat{z}}}(\hat{y}_{i}|\bm{\hat{z}})=(\mathcal{N}(\mu_{i},\sigma_{i}^2)\ast\mathcal{U}(-\frac{1}{2},\frac{1}{2}))(\hat{y_{i}})\\
  &=\frac{1}{2}erfc(-\frac{\hat{y_{i}}+\frac{1}{2}-\mu_{i}}{\sqrt{2}\sigma_{i}})-\frac{1}{2}erfc(-\frac{\hat{y_{i}}-\frac{1}{2}-\mu_{i}}{\sqrt{2}\sigma_{i}}).
\end{split}
\end{equation}
In (3), $\mathcal{U}$ is the uniform distribution for reproducing quantisation noise and $erfc$ is the complementary error function.
This equation can also be expressed using the cumulative distribution function $c$ of the Gaussian distribution as follows:
\begin{equation}
    p_{\bm{\hat{y}}|\bm{\hat{z}}}(\hat{y}_{i}|\bm{\hat{z}})=c(\hat{y_{i}}+\frac{1}{2})-c(\hat{y_{i}}-\frac{1}{2}).
\end{equation}

Several other entropy models have been proposed as encoding methods to be incorporated into LIC for human vision \cite{a22,a23}.
In the study by Cheng \textit{et al}.\cite{a23}, a mixed normal distribution is assumed as the probability distribution of the feature $\bm{y}$, that is,
\begin{equation}
  p_{\bm{\hat{y}}|\bm{\hat{z}}}(\bm{\hat{y}}|\bm{\hat{z}})\sim\sum_{k=1}^{K}\bm{w}^{(k)}\mathcal{N}(\bm{\mu}^{(k)},\bm{\sigma}^{2(k)}).
\end{equation}
Fu \textit{et al}.\cite{a22} have proposed an entropy model using Laplacian and Logistic distributions.
These innovative designs in entropy models have contributed to improving the compression performance targeting human vision.
LIC models with these entropy models are optimized using the following loss functions:
\begin{equation}
  \mathcal{L}=\mathcal{R}(\bm{y})+\mathcal{R}(\bm{z})+\lambda \cdot mse(\bm{x},\bm{\hat{x}}).
\end{equation}
In (6), $\mathcal{R}(\bm{y})$ and $\mathcal{R}(\bm{z})$ are the bitrates of $\bm{y}$ and $\bm{z}$, respectively.
$\bm{x}$ represents the input image, and $\bm{\hat{x}}$ represents the decoder output image.
$mse$ represents the mean squared error function and $\lambda$ is a constant to control the rate.

The design of the entropy model and loss function in LIC has been extensively studied as an image compression method for human vision.
This design has also been integrated into most ICM methods, achieving some performance as an image compression method aimed at machine vision.
However, we believe that the design of LIC in ICM is open to review.

\begin{figure*}[bt]
    \centerline{\includegraphics[width=2\columnwidth]{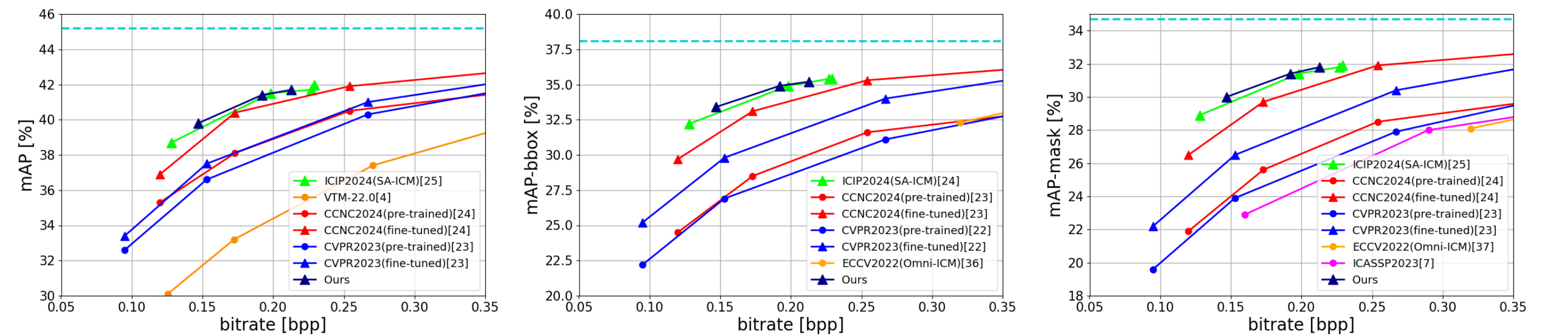}}
    \caption{Image compression performance for image recognition tasks. (Left): Object detection accuracy with Yolov5 versus bitrate; (Middle): Object detection accuracy with Mask R-CNN versus bitrate; (Right): Segmentation accuracy with Mask R-CNN versus bitrate.}
    \label{fig:graph}
    \end{figure*}

\subsection{Image Coding for Machines}
As image recognition accuracy improves, image compression techniques for machines have become essential to make image recognition models more prevalent.
This technique is known as "Image Coding for Machines (ICM)," and many methods have been proposed. 
Most of these methods involve modifications to the LIC model used for human vision. 
There are two main types of modifications in ICM.

The first modification method is to increase the input to the LIC model.
In this approach, an ROI-map is provided to the LIC model along with the image to be compressed \cite{a24,a25}.
This ensures that more bits are allocated to parts of the image needed for image recognition.
The advantage of this method is its broad applicability to object detection and instance segmentation tasts.
There are also methods to incorporate ROI-map into pre-trained LIC models or rule-based codecs like VVC and HEVC \cite{a5,a26}.
These methods do not require additional training and can be integrated into existing compression methods.
However, using the ROI-map as input to the LIC model means that the ROI-map must be estimated before compressing the image. 

The second method for modification involves changing the loss function used to train the LIC model.
A new loss component is added to the loss function of the LIC model shown in (6).
In the ICM method, there are two main additional loss components: task-loss and region-learning-loss.
Task-loss measures the error between the image recognition result obtained by inputting the decoded image into the recognition model and the ground truth label \cite{a8,a9,a10}.
Training the LIC model to minimize this loss makes it possible to decode images for machines while reducing the bitrate.
The loss function of the LIC model, including task loss, is expressed by the following equation.
\begin{equation}
  \mathcal{L}=\mathcal{R}(\bm{y})+\mathcal{R}(\bm{z})+\lambda_{1} \cdot mse(\bm{x},\bm{\hat{x}})+\lambda_{2} \cdot \mathcal{M}(\bm{\hat{x}}).
\end{equation}
In (7), $\mathcal{R}$, $mse$, $\bm{y}$, $\bm{z}$, $\bm{x}$, and $\bm{\hat{x}}$ have the same meaning as those functions, variables, and constants in (6).
$\mathcal{M}(\bm{\hat{x}})$ is the task-loss that can be computed by inputting the coded image into the image recognition model.
$\lambda_{1}$ and $\lambda_{2}$ are constants to control the rate.
ICM methods using task loss can optimize the LIC model for image recognition.
On the other hand, these compression methods can only achieve image compression for the specific image recognition model used during their optimization.

Region-learning-loss is the MSE between the decoded image and the original image in a specific part of the image \cite{a20,a21}.
To learn how to decode a specific part of the image, a mask image is needed to indicate that part during training.
The loss function of the LIC model with region-learning-loss is expressed by the following equation:
\begin{equation}
  \mathcal{L}=\mathcal{R}(\bm{y})+\mathcal{R}(\bm{z})+\lambda \cdot mse(\bm{x} \odot m_{x},\bm{\hat{x}} \odot m_{x}).
\end{equation}
In (8), $m_{x}$ is the binary mask image corresponding to input image $\bm{x}$.
Other variables and functions have the same meaning as those shown in (6).
As mask images, some methods use handcrafted segmentation labels or segmentation results from the Segment Anything Model (SAM) \cite{a27}.
Specifically, SA-ICM enables the LIC to learn a compression method for edge regions created using SAM, as shown in Fig.\ref{fig:saicm}.
The mask image is only needed during the training of the LIC model, not during testing.
The advantage of this learning method is that it can decode suitable images for various image recognition tasks without requiring additional information such as ROI-maps or mask images.
In addition, this approach can be extended to image coding methods for humans and machines \cite{a28,a29}.
In this paper, the ICM method with region-learning-loss is used to investigate better-performing image compression methods.

\section{Proposed Method}
\label{sec:method}

\subsection{Entropy Model with Delta Function}
We propose a new ICM method called Delta-ICM, which incorporates a delta function into the entropy model.
Typically, most LIC models used for image compression for both humans and machines utilize the entropy model shown in (3).
This entropy model is effective when the entire image needs to be decoded.
However, as we can see from the loss function (8), the ICM method does not always require the entire image to be decoded, allowing us to reconsider the design of this entropy model.
To efficiently eliminate unnecessary information from parts of the image that do not need to be decoded, we introduce a delta function into the entropy model.
This entropy model ensures that a feature with a specific value has a probability of 1, effectively reducing the information content of that feature.
The delta function can be treated like a probability distribution because its integral from $-\infty$ to $\infty$ equals $1$. 
Therefore, in this paper, the probability distribution represented by the delta function $\delta(x-a)$ is referred to as a delta distribution and is denoted as $\delta(a)$. 
By incorporating this distribution into the entropy model, we attempt to efficiently eliminate the parts of the image that are unnecessary for recognition.
Assuming that the latent feature $y$ of the image follows only a delta distribution under the conditioning by feature $z$, the entropy model can be expressed as follows:
\begin{equation}
  \begin{split}
    p_{\bm{\hat{y}}|\bm{\hat{z}}}(\hat{y}_{i}|\bm{\hat{z}})&=(\delta(\mu_{i})\ast\mathcal{U}(-\frac{1}{2},\frac{1}{2}))(\hat{y_{i}})\\
    &=F(\hat{y}_{i}+\frac{1}{2}-\mu_{i})-F(\hat{y}_{i}-\frac{1}{2}-\mu_{i}).
  \end{split}
\end{equation}
In (9), $F$ is the cumulative distribution function of the delta distribution.
This can be expressed as following:
\begin{equation}
  F(\xi)=\int_{-\infty}^{\xi}\delta(t)dt=
  \left\{ 
  \begin{array}{ll}
    1 & (\xi \geq 0) \\
    0 & (\xi < 0)
\end{array} \right..
\end{equation}

In the ICM method, the parts of the image needed for machines must be decoded.
For these parts, we aim to use the entropy model with the normal distribution, as shown in (2).
Therefore, we propose a method where either the delta distribution or the normal distribution is selected for each image part, and the appropriate distribution is used to encode the parts.
In the proposed method, the latent features $\bm{y}$ of the image are assumed to follow a mixture of normal and delta distributions, that is,
\begin{equation}
  p_{\bm{\hat{y}}|\bm{\hat{z}}}(\bm{\hat{y}}|\bm{\hat{z}})\sim\{\bm{w}\mathcal{N}(\bm{\mu},\bm{\sigma}^2)+(1-\bm{w})\delta(\bm{\mu})\}.
\end{equation}
In (11), $\bm{w}$ is the parameter to be optimized in training the LIC model.
This parameter balances the application of the delta distribution with that of the normal distribution.
The entropy model is also expressed by the following equation:
\begin{equation}
  \begin{split}
    &p_{\bm{\hat{y}}|\bm{\hat{z}}}(\hat{y}_{i}|\bm{\hat{z}})=\\
    &(\{w_{i}\mathcal{N}(\mu_{i},\sigma_{i}^2)+(1-w_{i})\delta(\mu_{i})\}\ast\mathcal{U}(-\frac{1}{2},\frac{1}{2}))(\hat{y_{i}}).
  \end{split}
\end{equation}
The training of the LIC model with this entropy model enables the encoding of latent features using both delta and normal distributions.

\begin{table}[t]
    \centering
    \caption{Performance comparison between SA-ICm and Delta-ICM.} \label{tab:comp}
    \small
    \begin{tabular}{c|c|c}

      \hline
      \multicolumn{3}{c}{BD-mAP [\%]} \\
      \cline{1-3}
      Yolov5    & Mask R-CNN (det.) & Mask R-CNN (seg.)\\

      \hline
      \hline
      -3.22 \%  & -4.50 \% & -4.92 \%\\
      \hline
    \end{tabular}
  \end{table}
\begin{figure}[bt]
    \centerline{\includegraphics[width=1\columnwidth]{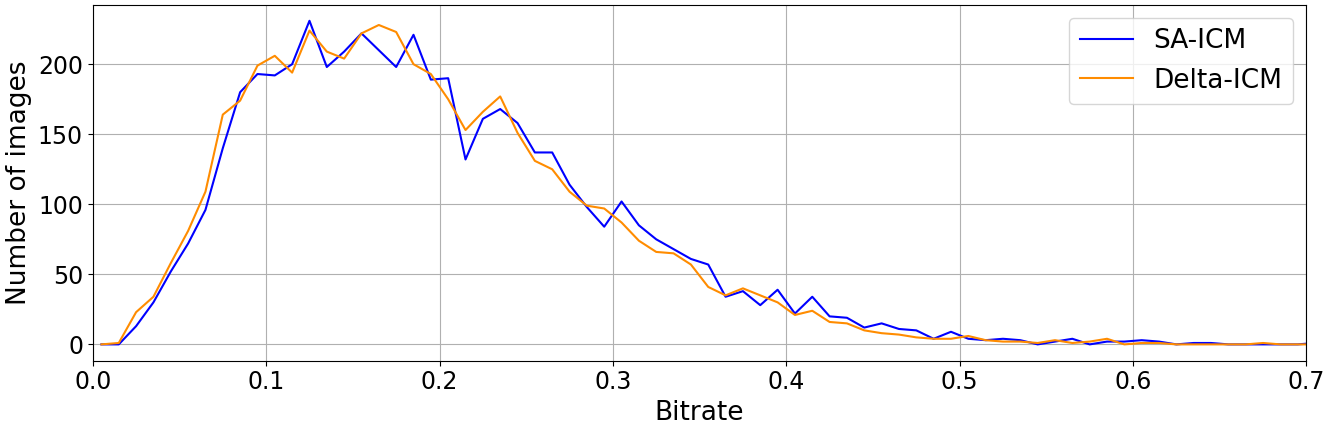}}
    \caption{Bitrate of decoded images in the COCO (validation) dataset using Delta-ICM and SA-ICM, respectively. The only difference between these ICM methods is the entropy model design method. All other design and training methods are the same.}
    \label{fig:comp}
    \end{figure}

\section{Experiment}
\subsection{Image Coding for Yolov5}
To validate the effectiveness of the proposed method, we evaluate its image compression performance on machine tasks. 
In the first experiment, we use Yolov5 \cite{a30} as the image recognition model for an object detection task. 
COCO (train) dataset \cite{a31} is used to train Delta-ICM and (8) is used as the loss function. 
Delta-ICM is trained using 0.03, 0.05, and 0.06 as the values of $\lambda$ in (8). 
After training Delta-ICM, we compress the COCO (train) and COCO (validation) datasets to generate the corresponding coded image groups.
An example of coded images is shown in Fig.\ref{fig:exp}.
These image groups are then used for training and testing Yolov5.

The image compression performance for Yolov5 is shown in Fig.\ref{fig:graph}. 
The vertical axis represents the object detection accuracy (mAP), while the horizontal axis represents the bitrate. 
Navy curves represent the image compression performance of the proposed method, while the other curves represent the performance of the comparison method. 
In particular, the curves in red and green are the conventional ICM methods \cite{a20,a21}. 
From this graph, the proposed method shows better compression performance than the comparison method.

\begin{figure}[t]
    \centerline{\includegraphics[width=1\columnwidth]{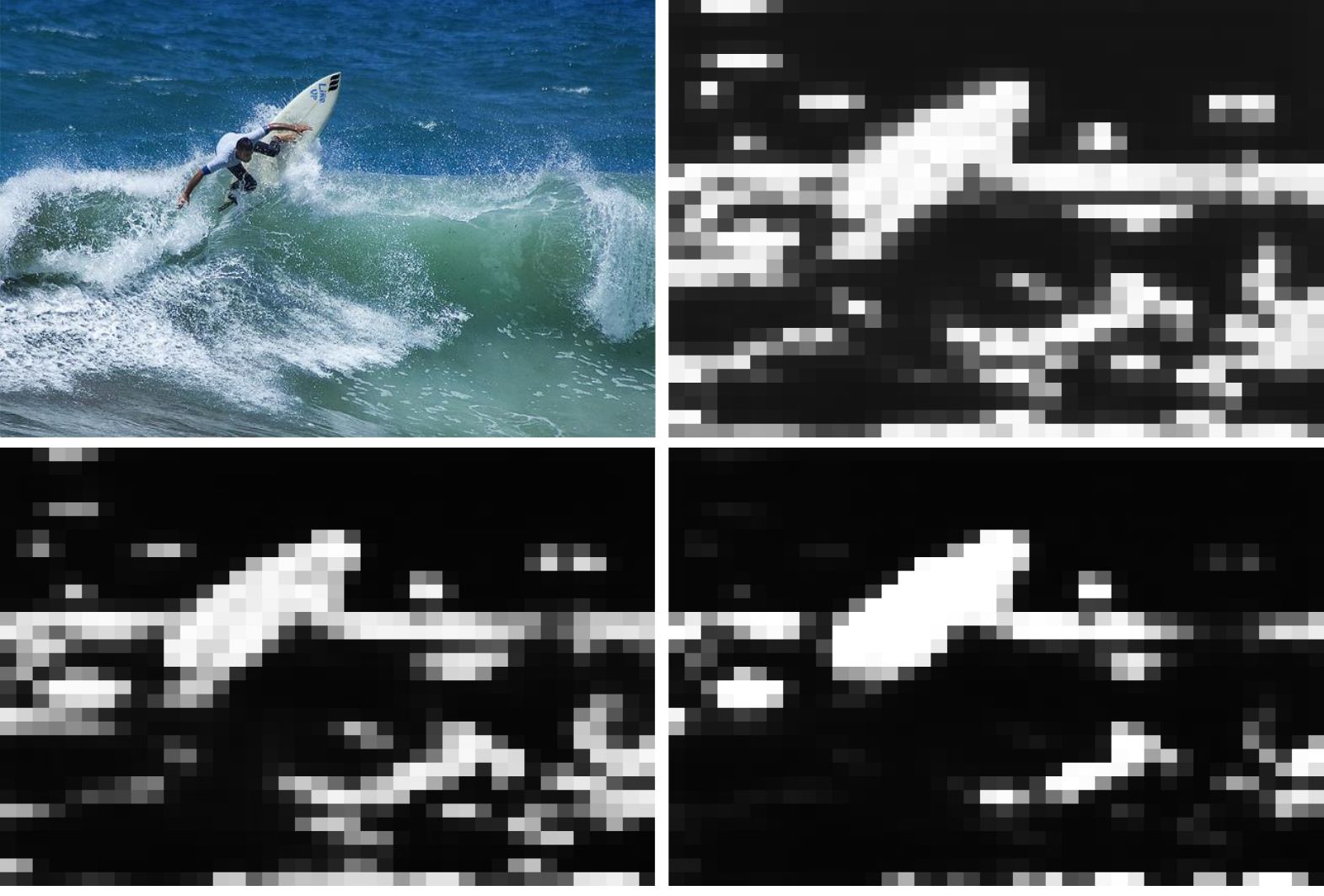}}
    \caption{Ratio of Gaussian to Delta distribution in Delta-ICM. The grayscale images represent the weights in (11). Black areas indicate the use of the delta distribution, while white areas correspond to the Gaussian distribution.}
    \label{fig:weight}
    \end{figure}
    
\subsection{Image Coding for Mask R-CNN}
In the second experiment, Mask R-CNN \cite{a32} is used as the image recognition model for both object detection and segmentation tasks. 
As in the first experiment, Delta-ICM is employed to generate coded images from the COCO (train) and COCO (validation) datasets. 
These coded image groups are used for training and testing Mask R-CNN, respectively. The image compression performance for Mask R-CNN is shown in Fig.\ref{fig:graph}. 
As in Fig.\ref{fig:graph}, the Navy curve shows the compression performance of the proposed method. The other colors show the compression performance of the comparison method. 
In particular, the red, green, pink and orange curves are the conventional ICM method \cite{a20,a21,a7,a33}. 
From these graphs, it is evident that the proposed method achieves superior image compression performance for image recognition compared to the conventional ICM method.

\subsection{Comparison of Delta-ICM and SA-ICM}
The proposed method and SA-ICM share almost the same model structure and use the same loss function for training. 
The key difference between these methods lies in the entropy model. 
The entropy model of Delta-ICM is constructed using both delta and normal distributions, as described in (11), while the entropy model of SA-ICM uses only the normal distribution, as in (2). 
To validate the design of our entropy model, we compare the performance of Delta-ICM and SA-ICM.
BD-mAP[\%] is calculated for each of the three graphs in the Fig.\ref{fig:graph}, and the results are shown in Table \ref{tab:comp}.
This table shows that Delta-ICM can achieve the same recognition accuracy with 3,4[\%] less bitrate than SA-ICM.
Fig.\ref{fig:comp} presents a comparison between Delta-ICM and SA-ICM using the exact same values for $\lambda (0.05)$ and $m$ in (8). 
This graph shows the distribution of the bitrate for the COCO (validation) dataset, compressed using these two ICM methods. 
The vertical axis indicates the number of images and the horizontal axis indicates the bitrate. 
It can be seen that the distribution of the bitrate of the compressed image by Delta-ICM is to the left of that of SA-ICM.
This confirms the bitrate reduction due to the introduction of the delta function to the entropy modeling.

Additionally, in Fig.\ref{fig:weight}, we visualize the value of the variable $\bm{w}$ from (11).
The white and black regions in the figure indicate areas coded using the normal and delta distributions, respectively. 
This visualization confirms that our entropy model design effectively separates the distribution used, depending on the different parts of the image.

\section{Conclusion}
We proposed an entropy model for learned image compression using the delta function.
In designing the entropy model, we utilized both gaussian and delta distributions, allowing us to choose the optimal distribution for each image feature.
By applying this entropy model to the ICM method, our experiments demonstrated that this model effectively discards unnecessary image textures for image recognition.
We also validated the performance of Delta-ICM by comparing its compression efficiency to conventional ICM methods.
Using Yolov5 and Mask R-CNN as image recognition models, and object detection and segmentation as tasks, we confirmed the suitability of our proposed method for various models and tasks.
Future research should focus on developing an improved entropy model to further enhance compression performance.
Additionally, the method for reducing the model's weight to integrate ICM technology into smart devices should be discussed.

\section*{Acknowledgment}
The results of this research were obtained from the commissioned research (JPJ012368C05101) by National Institute of Information and Communications Technology (NICT), Japan.

\vspace{12pt}


\begin{thebibliography}{00}
  \bibitem{a1} G. K. Wallace, ``The JPEG still picture compression standard,'' IEEE Transactions on Consumer Electronics, vol. 38, no. 1, pp. xviii-xxxiv, Feb. 1992.
  \bibitem{a2} ITU-T and ISO/IEC JTC 1, Advanced video coding for generic audiovisual services, ITU-T Recommendation H.264 and ISO/IEC 14496-10 (MPEG-4 AVC), 2010.
  \bibitem{a3} High Efficiency Video Coding, Standard ISO/IEC 23008-2, ISO/IEC JTC 1, Apr. 2013.
  \bibitem{a4} Versatile Video Coding, Standard ISO/IEC 23090-3, ISO/IEC JTC 1, Jul. 2020.
  \bibitem{a5} H. Choi and I. V. Bajic, "High Efficiency Compression for Object Detection," 2018 IEEE International Conference on Acoustics, Speech and Signal Processing (ICASSP), 2018, pp. 1792-1796.
  \bibitem{a6} Z. Huang, C. Jia, S. Wang and S. Ma, "Visual Analysis Motivated Rate-Distortion Model for Image Coding," 2021 IEEE International Conference on Multimedia and Expo (ICME), 2021, pp. 1-6.
  \bibitem{a7} B. Li, J. Liang, H. Fu and J. Han, "ROI-Based Deep Image Compression with Swin Transformers," ICASSP 2023 - 2023 IEEE International Conference on Acoustics, Speech and Signal Processing (ICASSP), 2023, pp. 1-5.
  \bibitem{c1} T. Shindo, T. Watanabe, K. Yamada and H. Watanabe, "VVC Extension Scheme for Object Detection Using Contrast Reduction," 2023 IEEE 12th Global Conference on Consumer Electronics (GCCE), 2023, pp. 1097-1098.
  \bibitem{b3} T. Shindo, T. Watanabe, K. Yamada and H. Watanabe, "Accuracy Improvement of Object Detection in VVC Coded Video Using YOLO-v7 Features," 2023 IEEE International Conference on Artificial Intelligence in Engineering and Technology, 2023, pp. 247-251.
  \bibitem{a8} F. Codevilla, J. G. Simard, R. Goroshin, and C. Pal, “Learned Image Compression for Machine Perception,” arXiv prepint, arXiv : 2111.02249, 2021.
  \bibitem{a9} N. Le, H. Zhang, F. Cricri, R. Ghaznavi-Youvalari, and E. Rahtu, "Image Coding For Machines: an End-To-End Learned Approach," ICASSP 2021 - 2021 IEEE International Conference on Acoustics, Speech and Signal Processing (ICASSP), 2021, pp. 1590-1594.
  \bibitem{a10} N. Le, H. Zhang, F. Cricri, R. Ghaznavi-Youvalari, H. R. Tavakoli and E. Rahtu, "Learned Image Coding for Machines: A Content-Adaptive Approach," 2021 IEEE International Conference on Multimedia and Expo (ICME), 2021, pp. 1-6.
  \bibitem{b2} B. Azizian and I. V. Bajić, "Privacy-Preserving Feature Coding for Machines," 2022 Picture Coding Symposium (PCS), 2022, pp. 205-209.
  \bibitem{a11} J. Ballé, V. Laparra and E. P. Simoncelli, “End-to-end Optimized Image Compression,” in Proc. International Conference on Learning Representations (ICLR), Apr. 2017, pp. 1–27.
  \bibitem{a12} D. Minnen and S. Singh, "Channel-Wise Autoregressive Entropy Models for Learned Image Compression," 2020 IEEE International Conference on Image Processing (ICIP), 2020, pp. 3339-3343.
  \bibitem{a13} D. Minnen, G. Toderici, S. Singh, S. J. Hwang and M. Covell, "Image-Dependent Local Entropy Models for Learned Image Compression," 2018 25th IEEE International Conference on Image Processing (ICIP), Athens, Greece, 2018, pp. 430-434.
  \bibitem{b1} D. He, Y. Zheng, B. Sun, Y. Wang and H. Qin, "Checkerboard Context Model for Efficient Learned Image Compression," 2021 IEEE/CVF Conference on Computer Vision and Pattern Recognition (CVPR), Nashville, TN, USA, 2021, pp. 14766-14775.
  \bibitem{a14} J. Ballé, D. Minnen, S. Singh, S. J. Hwang and N. Johnston, “Variational image compression with a scale hyperprior,” in Proc. International Conference on Learning Representations (ICLR), May 2018, pp. 1–10.
  \bibitem{a15} D. Minnen, J. Ballé and G. Toderici, “Joint autoregressive and hierarchical priors for learned image compression." Advances in neural information processing systems 31 (2018).
  \bibitem{a16} J. Kim, B. Heo and J. Lee, "Joint Global and Local Hierarchical Priors for Learned Image Compression," 2022 IEEE/CVF Conference on Computer Vision and Pattern Recognition, 2022, pp. 5982-5991.
  \bibitem{a17} D. He, Z. Yang, W. Peng, R. Ma, H. Qin and Y. Wang, "ELIC: Efficient Learned Image Compression with Unevenly Grouped Space-Channel Contextual Adaptive Coding," 2022 IEEE/CVF Conference on Computer Vision and Pattern Recognition (CVPR), 2022, pp. 5708-5717.
  \bibitem{a18} W. Duan et al., "Learned Image Compression Using Cross-Component Attention Mechanism," in IEEE Transactions on Image Processing, vol. 32, pp. 5478-5493, 2023.
  \bibitem{a19} J. Liu, H. Sun and J. Katto, "Learned Image Compression with Mixed Transformer-CNN Architectures," 2023 IEEE/CVF Conference on Computer Vision and Pattern Recognition (CVPR), Vancouver, BC, Canada, 2023, pp. 14388-14397.
  \bibitem{a20} T. Shindo, T. Watanabe, K. Yamada and H. Watanabe, "Image Coding for Machines with Object Region Learning," 2024 IEEE 21st Consumer Communications and Networking Conference (CCNC), Las Vegas, NV, USA, 2024, pp. 1040-1041.
  \bibitem{a21} T. Shindo, K. Yamada, T. Watanabe and H. Watanabe, "Image Coding For Machines With Edge Information Learning Using Segment Anything," 2024 IEEE International Conference on Image Processing (ICIP), Abu Dhabi, United Arab Emirates, 2024, pp. 3702-3708.
  \bibitem{a22} H. Fu et al., "Learned Image Compression With Gaussian-Laplacian-Logistic Mixture Model and Concatenated Residual Modules," in IEEE Transactions on Image Processing, vol. 32, pp. 2063-2076, 2023.
  \bibitem{a23} Z. Cheng, H. Sun, M. Takeuchi and J. Katto, "Learned Image Compression With Discretized Gaussian Mixture Likelihoods and Attention Modules," 2020 IEEE/CVF Conference on Computer Vision and Pattern Recognition (CVPR), Seattle, WA, USA, 2020, pp. 7936-7945.
  \bibitem{a24} N. Le et al., "Bridging the Gap Between Image Coding for Machines and Humans," 2022 IEEE International Conference on Image Processing (ICIP), Bordeaux, France, 2022, pp. 3411-3415.
  \bibitem{a25} J. I. Ahonen, N. Le, H. Zhang, F. Cricri and E. Rahtu, "Region of Interest Enabled Learned Image Coding for Machines," 2023 IEEE 25th International Workshop on Multimedia Signal Processing (MMSP), Poitiers, France, 2023, pp. 1-6.
  \bibitem{a26} J. I. Ahonen et al., "NN-VVC: Versatile Video Coding boosted by self-supervisedly learned image coding for machines," 2023 IEEE International Symposium on Multimedia (ISM), 2023, pp. 10-19.
  \bibitem{a27} A. Kirillov \textit{et al}., “Segment Anything,” Proceedings of the IEEE/CVF International Conference on Computer Vision (ICCV), 2023, pp. 4015-4026.
  \bibitem{a28} T. Shindo, T. Watanabe, Y. Tatsumi and H. Watanabe, "Scalable Image Coding for Humans and Machines Using Feature Fusion Network," arXiv e-prints, 2024,	arXiv:2405.09152.
  \bibitem{a29} T. Shindo, Y. Tatsumi, T. Watanabe and H. Watanabe, "Refining Coded Image in Human Vision Layer Using CNN-Based Post-Processing," arXiv e-prints, 2024,	arXiv:2405.11894.
  
  \bibitem{a30} G. Jocher \textit{et al}., "ultralytics/yolov5: v7.0-yolov5 sota realtime instance segmentation," Zenodo, Nov., 2022.
  \bibitem{a31} T. Y. Lin \textit{et al}., "Microsoft COCO: Common Objects in Context," Computer Vision - ECCV 2014. ECCV 2014. Lecture Notes in Computer Science, vol 8693. 2014, pp 740-755.
  \bibitem{a32} K. He, G. Gkioxari, P. Dollar, and R. Girshick, "Mask r-cnn," Proceedings of the IEEE international conference on computer vision (ICCV), 2017, pp. 2961-2969.
  \bibitem{a33} R. Feng, X. Jin, Z. Guo, R. Feng, Y. Gao, T. He, Z. Zhang, S. Sun, and Z. Chen, “Image Coding for Machines with Omnipotent Feature Learning,” Computer Vision - ECCV 2022. ECCV 2022. Lecture Notes in Computer Science, vol 13697. 2022, pp 510-528.
  
\end{thebibliography}
\end{document}